\documentclass{article}

\usepackage[final]{corl_2020} % Uncomment for the camera-ready ``final'' version.

\usepackage{color}
\usepackage{epsfig}
\usepackage{graphicx}
\usepackage{algorithm,algorithmic}
\usepackage{adjustbox}
\usepackage{array}
\usepackage{booktabs}
\usepackage{colortbl}
\usepackage{float,wrapfig}
\usepackage{framed}
\usepackage{hhline}
\usepackage{multirow}
\usepackage{subcaption} % issues a warning with CVPR/ICCV format
\usepackage[font=small]{caption}
\usepackage[percent]{overpic}
\usepackage{amsmath,amsfonts,amssymb}
\usepackage{amsthm} 
\usepackage{bm}
\usepackage{nicefrac}
\usepackage{microtype}
\usepackage{contour}
\usepackage{courier}
\usepackage{changepage}
\usepackage{extramarks}
\usepackage{fancyhdr}
\usepackage{lastpage}
\usepackage{setspace}
\usepackage{soul}
\usepackage{xspace}
\usepackage{cuted}
\usepackage{fancybox}
\usepackage{afterpage}
\usepackage{url}
\usepackage{quoting}
\usepackage{epigraph}

\usepackage{enumerate}
\usepackage{paralist,tabularx}
\usepackage{comment}
\usepackage{pdfpages}
\DeclareMathOperator*{\argmin}{arg\,min}
\DeclareMathOperator*{\argmax}{arg\,max}

\makeatletter
\DeclareRobustCommand\onedot{\futurelet\@let@token\@onedot}
\def\@onedot{\ifx\@let@token.\else.\null\fi\xspace}

\def\eg{e.g\onedot} 
\def\ie{i.e\onedot}

\def\etal{et al\onedot}
\makeatother

\definecolor{MyDarkBlue}{rgb}{0,0.08,1}
\definecolor{MyDarkGreen}{rgb}{0.02,0.6,0.02}
\definecolor{MyDarkRed}{rgb}{0.8,0.02,0.02}
\definecolor{MyDarkOrange}{rgb}{0.40,0.2,0.02}
\definecolor{MyPurple}{RGB}{111,0,255}
\definecolor{MyRed}{rgb}{1.0,0.0,0.0}
\definecolor{MyGold}{rgb}{0.75,0.6,0.12}
\definecolor{MyDarkgray}{rgb}{0.66, 0.66, 0.66}

\newcommand{\shuran}[1]{\textcolor{MyDarkGreen}{[Shuran: #1]}}

\newcommand{\jw}[1]{\textcolor{MyRed}{[Jiajun: #1]}}

\newcommand{\mypara}{\vspace{-2.5mm}\paragraph}

\def\OURSFull{3D Dynamic Scene Representation\xspace}
\def\OURS{DSR-Net\xspace}

\def\BaselinePixelFlow{2DFlow\xspace}
\def\BaselineSE{SE3-Net\xspace}
\def\BaselineSEPose{SE3Pose-Net\xspace}
\def\BaselineVoxelFlow{3DFlow\xspace}
\def\BaselineSingle{SingleStep\xspace}
\def\BaselineNoWarp{NoWarp\xspace}
\def\BaselineGTWarp{GTWarp\xspace}

\newcommand\blfootnote[1]{%
  \begingroup
  \renewcommand\thefootnote{}\footnote{#1}%
  \addtocounter{footnote}{-1}%
  \endgroup
}

\title{Learning 3D Dynamic Scene Representations\\ for Robot Manipulation}

\author{
  Zhenjia Xu$^{1*}$, Zhanpeng He$^{1*}$, Jiajun Wu$^{2}$, Shuran Song$^{1}$\\
  $^{1}$ Columbia University, $^{2}$ Stanford University\\
  \url{https://dsr-net.cs.columbia.edu/}
}

\begin{document}
\maketitle

%===============================================================================

\blfootnote{$*$ Indicates equal contribution}
\vspace{-10mm}
\begin{abstract}
%%%ALL don't use marcos in abstract - hard to copy paste 
3D scene representation for robot manipulation should capture three key object properties: permanency -- objects that become occluded over time continue to exist; amodal completeness -- objects have 3D occupancy, even if only partial observations are available; spatiotemporal continuity -- the movement of each object is continuous over space and time. In this paper, we introduce 3D Dynamic Scene Representation (DSR), a 3D volumetric scene representation that simultaneously discovers, tracks, reconstructs objects, and predicts their dynamics while capturing all three properties. We further propose DSR-Net, which learns to aggregate visual observations over multiple interactions to gradually build and refine DSR. Our model achieves state-of-the-art performance in modeling 3D scene dynamics with DSR on both simulated and real data. Combined with model predictive control, DSR-Net enables accurate planning in downstream robotic manipulation tasks such as planar pushing. Code and data are available at \url{dsr-net.cs.columbia.edu}.

%Code and data will be available online. 

% DSR models the 3D world in a way that is persistent, temporally consistent, and interpretable -- properties that are desirable as part of a perception stack for downstream robotics manipulation tasks. %\jw{I suggest we define the three properties, especially "persistent". I don't think it's a concept well-known to the community.} 
%to which predicts objects' 3D motion (encoded as 3D scene flow) conditioned on visual observation and action. 
%\jw{which predicts ***. this doesn't sound novel. What's new in our model? Temporal aggregation? If so, it'd be better to highlight that in this first sentence.} 
%Our model learns to aggregate information of multiple time steps to obtain a representation with desirable properties we mention. \jw{Again, I feel the properties are not well defined, and it's unclear why they are desirable.} 

%\jw{I like the abstract. On the other hand, if we'd like to highlight the three (or four? + amodal as mentioned in the intro), we may want to rewrite it as:  "A desirable 3D scene representation should have four properties: 3D amodal complete, ****; persistent, so that ***; temporally consistent, ***; and interpretable, ***. Current models lack in these aspects. We introduce 3D *** (DSR), which ***." and some sentences in the second half of the current abstract may be shortened or merged.} \shuran{agree, please go ahead and change them }
\end{abstract}

\iffalse
We introduce \OURSFull (\OURS), a deep volumetric representation that enables simultaneous discovery, tracking, and reconstruction of novel objects and prediction of their motion under robot interaction. %a deep volumetric representation that  as  it  continually aggregates new observations from depth images.
\jw{maybe add a sentence to explain what DSR is.}
This representation models the 3D world in a way that is persistent, temporally consistent, and interpretable -- properties that are desirable as part of a perception stack for downstream robotics tasks such as manipulation.
\jw{this seems to be from nowhere. Maybe remove the previous sentence.}
We propose DSR-Net, a 3D convolutional recurrent neural network that \jw{"convolutional recurrent" is awkward to me. Maybe "We propose DSR-Net, taking in visual ..."} takes in visual observations and actions, and outputs objects' motion, amodal representations of the current and next state. Our model utilizes information of multiple time steps and aggregates this information to a feature space that allows our model to learn a representation with desirable properties we mention. Our model achieves state-of-the-art performance in modeling 3D scene dynamics on both simulated and real data. Combined with model predictive control (MPC), \OURS enables accurate planning in downstream robotic manipulation tasks.
\fi 
\vspace{-2mm}

\vspace{-2mm}
\section{Introduction} \vspace{-2mm}
Our physical world is three-dimensional, where the full extent of objects -- their shape and motion -- exists and persists in 3D space. Despite this, the vast majority of visual predictive models currently used in robotics, which predict the motion of objects under the effect of an applied action, remain limited to only predicting 2D motion (\ie optical flow) of partial observations, \eg predicting the 2D flow of pixels \citep{vijayanarasimhan2017sfm,dosovitskiy2015flownet}, or predicting the 3D scene flow of visible points from a partial point cloud \citep{byravan2017se3,byravan2018se3}. Unfortunately, modeling the motion of only visible surfaces often leads to data degeneration, where objects fade and vanish from the representation as they become occluded. This causes the predictive models to perform poorly in cluttered environments, in which objects frequently appear, disappear, then reappear in view as the robot move them around.

In this work, we investigate the benefits of learning a complete and persistent 3D scene representation for visual predictive modeling. We present \textbf{\OURSFull (DSR)}: a 3D volumetric scene representation that simultaneously discovers, tracks and reconstructs novel objects and predicts their motion under a robot's interactions. %\jw{We need a sentence to explain what DSR is. It's never defined.}
%a deep volumetric representation that simultaneously \todo{discovers, tracks, and reconstructs novel objects on-the-fly as it continually aggregates new observations from depth images.} 
Specifically, the representation captures three object properties, all of which have long been argued as crucial to human scene understanding~\cite{spelke2007core}.
\begin{itemize}
    \vspace{-2mm}
    \item[\textbullet]  \textbf{{Permanence}:} visual information is aggregated into a persistent 3D representation. This means that as objects disappear from view due to occlusion, they remain in the representation. This enables more accurate predictions of object motion when it is moved by other objects in occlusion, or when it gradually reappears in view. %This is inspired by the phenomena of object permanence in human cognitive development.
    
    \item[\textbullet]   \textbf{Amodal completeness:} from partial observations of the scene, DSR infers complete 3D occupancy of each object, including regions that are not directly observed. This attribute enables it to predict the rigid body motion of the entire object instead of only visible surfaces.
    
    \item[\textbullet]   \textbf{Spatiotemporal continuity:} the representation recognizes individual object instances and tracks their identity over time.
    
    % \eg, as two bodies move independently, or are geometrically disjoint.
   \item[\textbullet]  \textbf{Interpretability:} DSR explicitly models object instances, geometry, and motion, makes it easy to be used out-of-the-box for high-level reasoning in robotic applications.
    \vspace{-2mm}
\end{itemize}

To learn this scene representation, we present \OURS, a 3D recurrent neural network that consists of two major components: 1) a scene encoder that encodes visual observations (\ie, depth images) into a volumetric 3D scene representation, and 2) a motion prediction network that takes in the 3D scene representation and an action to be performed by the robot and predicts volumetric scene flow. The scene flow is then used to spatially warp the current scene representation before combining it with the 3D scene representation of the next time step. The warping operation allows the network to aggregate information over time in a spatially coherent way. \OURS is trained end-to-end in simulation and then tested in the real world with a robotic manipulator on a tabletop setup. Our experiment result shows that our system achieves state-of-the-art performance in predicting the rigid body motion of novel objects under robot interaction in unstructured cluttered environments.

The contributions of our paper are three-fold.
First, we introduce a new 3D dynamic scene representation (DSR) that captures object permanence, amodal completeness, and continuity -- desirable properties as part of a perception stack for downstream robot manipulation tasks. 
Second, we propose \OURS, an end-to-end framework that learns such 3D representations via 3D convolutions. 
Third, we build a new benchmark dataset with over 80,000 simulated interactions and 1,500 real-world interactions for learning and evaluating dynamic 3D scene representations. 
Our experiments in both simulation and in the real-world show that \OURS achieves state-of-the-art performance in predicting 3D scene dynamics. Furthermore, it enables more accurate action planning in manipulation tasks such as planar pushing. Please find additional result and videos in supplementary material. 
%Our results also show that a key enabler in learning such representations that aggregate over time is the spatially coherent warping, producing high quality representations.

\begin{figure}[t]
    \centering
    \includegraphics[width=0.95\linewidth]{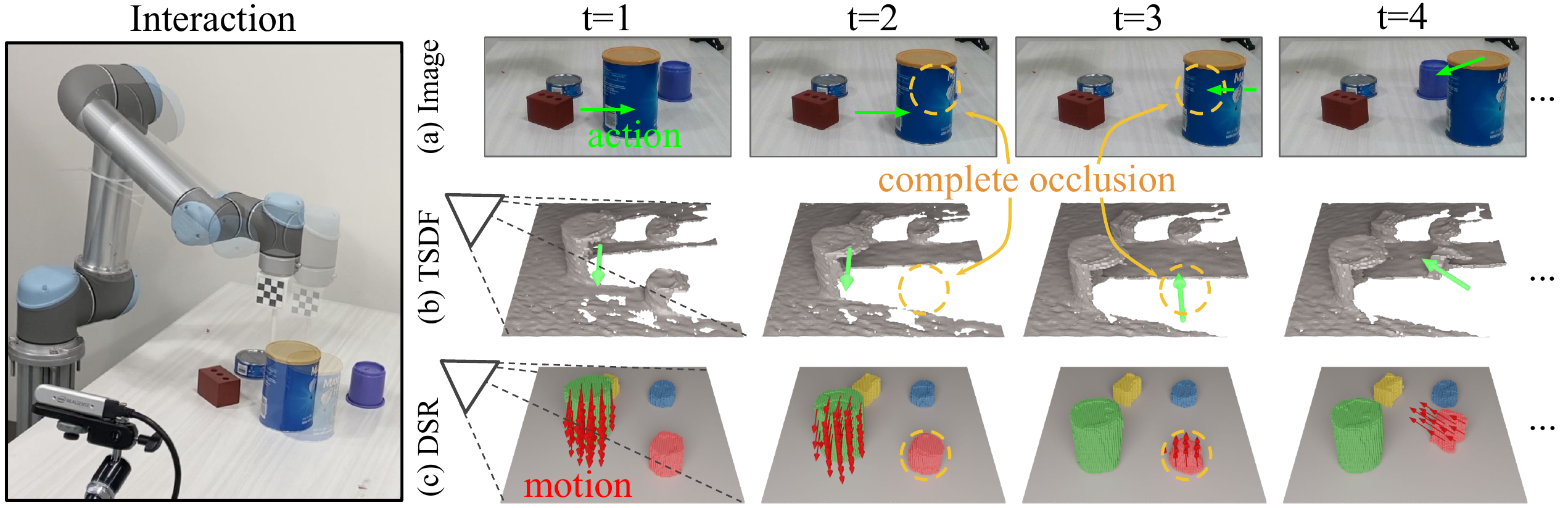}
    \caption{\textbf{Dynamic Scene Representation.} Given an applied action and a depth observation of a scene encoded as TSDF (b), the dynamic scene representation (c) is able to predict objects' motion (red arrows), infer the amodal 3D geometry of each object instance (colored voxels), and  maintain the object persistence under occlusion (t=2, object in orange circle). Color images (a) are used for visualization only. }
    \vspace{-5mm}
\end{figure}

% The goal of the work is to learn a 3D state representation that simultaneously discovers, tracks, and reconstructs novel objects on-the-fly as it continually aggregates new observations from a depth video with robots’ active interactions 
% The representation learned by the system features following four aspects: 
% Persistent:  Able to aggregate history
% Predictive: Able to 
% Amodal 3D: able to model the object using their complete 3D geometry instead of visible surface only
% Interpretable: Our representation explicitly models object instances and their  transformation, make it easier to use for other applications. 
% First two is unique for our representation that SE3 does not have 

% Key idea:  Better predictive model is enabled by \\
    % - complete 3D representation  - Make it easier to predict SE3 motion (need experiment) \\
	% - Persistent scene representation across time (don’t need to add or delete points) - make it easy for information aggregation\\
    % - Action centric motion representation \\
    
% We show the scene representation is useful  for robot motion planning 

\vspace{-2mm}
\section{Related work}
\vspace{-2mm}
Learning scene (or state) representations from visual data is a long-standing task in vision and robotics. 
Many different scene representations have been proposed for different environments, types of interaction, and applications. 
Our method learns an 3D scene representation for dynamic, multi-object  environments under robot interactions. 
Here we summarize most relevant works.

\mypara{Passive perception.} 
Most traditional computer vision tasks such as object detection or segmentation can be considered as extracting a high-level scene representation from passive observation, such as a single RGB image. 
However, these 2D representations cannot be directly applied in robotic applications that need to be operated in 3D. 
Recently many works have studied the problem of inferring 3D scene representations from partial observations such as a single color image ~\cite{garg2016unsupervised,zhan2018unsupervised} or a depth map~\cite{song2017semantic,prabhudesai2019embodied}. 
These representations are explicit, often in the form of 3D volumes or polygonal meshes. Latest papers in the field have also explored integrating neural nets for learning implicit 3D representations for objects and scenes~\cite{mescheder2019occupancy, chen2019learning, liu2019learning,jiang2020local}. 
While these scene representations have been used for robotics applications such as object grasping~\cite{van2019learning,varley2017shape}, they handle static environments only and cannot be directly applied in dynamic scenes. 

\mypara{Active perception.}
Systems that may update the camera viewpoint for exploration and representation building are often referred to as \textit{active perception} systems \cite{bajcsy1988active,bajcsy2018revisiting}. 
Cheng and Katerina~\cite{cheng2018geometry} proposed an active vision system that actively selects new camera viewpoints for estimating 3D object geometry and recognizing their identities. The representation learned by this system has been used for reinforcement learning~\cite{cheng2018reinforcement}. 
There are also models that actively learn a 3D scene representation from multi-view images or videos for better 3D geometry~\cite{kar2017learning,huang2018deepmvs}, shape correspondences~\cite{schmidt2016self,zeng20173dmatch,florence2018dense}, motion~\cite{harley2020learning,xie2019object}, semantic category~\cite{dai20183dmv, dai2018scancomplete}, or multiple objectives~\cite{vijayanarasimhan2017sfm}.
While active perception systems may collect additional information about object geometry with a moving camera, they still focus on static scenes, where there is no signal about object dynamics or physics.
In contrast, our model observes a robot's active interactions with objects in the scene (\eg, pushing). As a result, the scene representation can model and predict object dynamics under interaction, which is critical for task and action planning. %, where the goal is to actively change the object state towards the goal with interactions. 

\mypara{Interactive perception.}
Interactive perception is about perception facilitated by interaction with the environment~\cite{bohg2017interactive}. An important topic in interactive perception is on learning predictive and dynamic scene representations that are conditioned on current observations and interaction for manipulation\cite{agrawal2016learning,ebert2018visual}. Recently, a few visual predictive models have been proposed to learn an object-centric representation~\cite{ferreira2019learning,ye2019object,janner2018reasoning}, as well as to model 3D motion for rigid shapes~\cite{byravan2017se3,byravan2018se3}.
However, all these works predict motion in the form of per-pixel flow, which only considers the partial, observable surface, and does not leverage past interactions and observations. Therefore, the scene representations produced by these methods are often incomplete and inaccurate. The model that is most relevant to ours is DensePhysNet~\cite{xu2019densephysnet}, which learns to aggregate multiple interactions for a dense, 2D scene representation. However, it fails to model 3D relations, such as occlusion, and thus cannot provide a scene representation that maintains object permanence when there are occlusions. 

\vspace{-2mm}
\section{Dynamic Scene Representation Network (DSR-Net)}
\vspace{-2mm}
%We design Dynamic Scene Representation Network (DSR-Net) based on three insights. First, network should be able to explicitly model each object in scene and this scene representation should be consistent among different timesteps. This is critical for applying then learned representation to the down-stream manipulation tasks. Second, DSR-Net has a recurrent structure for information aggregation from multiple interactions to have a better scene representation. Third, information used for aggregation should take both scene and motion into consideration for a persistent representation to handle occlusion after the interaction.

\begin{figure}[t]
    \centering
    \includegraphics[width=\linewidth]{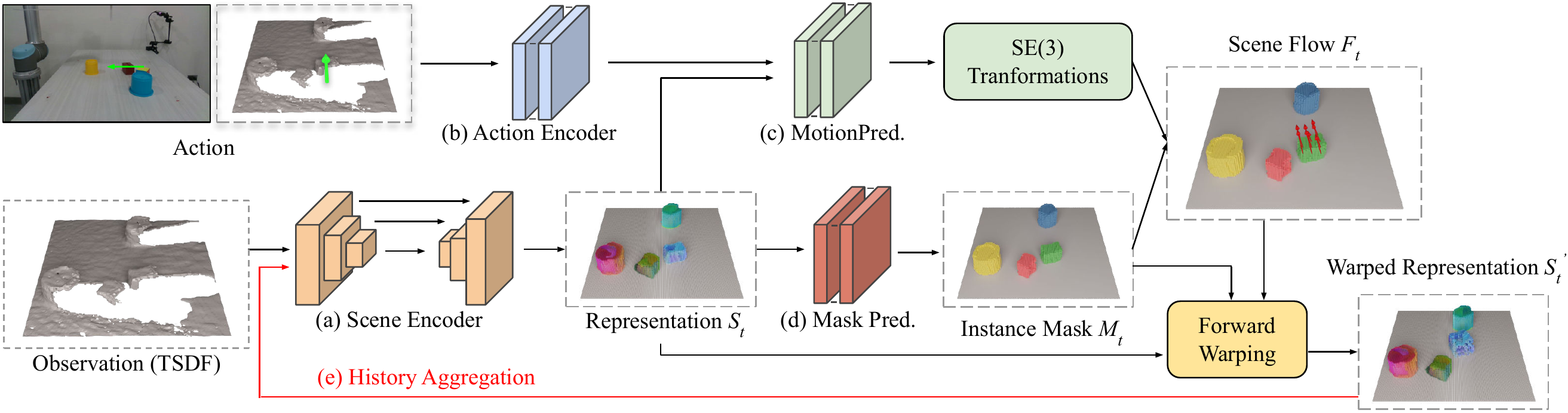}
    \caption{\textbf{DSR-Net}. 
    \OURS takes in the depth observation encoded as TSDF and action as input and predicts the amodal mask of each object $M_t$ and the voxel-wise scene flow $F_t$ after the interaction. The scene encoder (a) outputs a representation $S_t$ after aggregating the current observation and the history  ($S_t$ is colored by t-SNE embedding of the voxel-wise feature). $S_t$ is then used to predict (d) amodal object instance mask $M_t$. In parallel, the action encoder (b) encodes the input action and the motion predictor (c) predicts object motion represented as the SE(3) transformations. The scene flow $F_t$ is computed by combining the instance mask $M_t$ and transformations. The warped scene representation $S'_{t}$ is used as the history in the next step.}  %\jw{Warped representation is incorrectly labeled as $S'_t$ in the figure.} \jw{add a space between SceneEncoder (Scene Encoder), ActionEncoder, MotionPred., MaskPred.}}
    %https://docs.google.com/drawings/d/1pQNjexcv9Z-7YyY76JXCtNnUIOeVOe-Uji7Q0yvlAgE/edit?usp=sharing
    \label{fig:network}
    \vspace{-5mm}
\end{figure}

In this section, we first provide an overview of DSR-Net's network design and its advantages, then we provide description on each module and how to use it in robot manipulation.
Fig. \ref{fig:network} shows an overview of \OURS. At each step, the scene encoder outputs a 3D scene representation $S_t$ that aggregates the new observation (i.e., depth) with the past scene representation $S_{t-1}'$. $S_t$ is then used to predict amodal object instance mask $M_t$. In parallel, the motion predictor infers the object motion given the robot's action and current scene representation. The predicted motion and object instance mask are combined to compute voxel-level scene flow $F_t$. 
Finally, the scene flow $F_t$ is used to warp the scene representation $S_t$ to obtain $S'_{t}$ that is aligned with observation in the next interaction step and used for history aggregation. 

Our \OURS design provides four advantages compared to existing visual predictive models.
First, by using a 3D volumetric representation, \OURS naturally models objects' amodal 3D shape, regardless of occlusion. 
Second, by warping the previous scene representation using the predicted motion before concatenating it with the new observation, the network manages to aggregate history information in a spatially coherent manner: the same voxel stores information of the same object from past and new observations, regardless of their motion.  
Third, by leveraging history  information the representation is able to capture object permanence and continuity. 
Finally, all these properties allow the network to predict more accurate object motion and be useful for manipulation tasks.

%DSR-Net consists of two major components: 1) an encoder that takes in raw visual observations (in the form of depth images) and encodes them into a volumetric 3D scene representation, and a motion prediction network (Fig. \ref{fig:network} b) takes in the 3D scene representation and an action to be performed by the robot (\ie, a robot manipulator to execute a pushing action), and predicts volumetric scene flow. 
%The flow field is directly used to spatially warp the initial 3D scene representation (\ie, deep features) before combining it with the 3D scene representation of the next time step. This enables the network to aggregate information over time in a spatially coherent way. The DSR-Net is trained end-to-end in simulation, then tested in the real world with a robotic manipulator on a tabletop environment. This formulation enables our system to maintain a coherent 3D scene representation and a predictive model that achieves state-of-the-art performance in predicting the rigid body motion of novel objects in unstructured cluttered environments under the effects of a given robot action.

\mypara{Scene encoder (Fig. \ref{fig:network}a):}
Each depth observation is encoded with a truncated signed distance function (TSDF) with voxel size 0.004m.
The scene encoder concatenates the current observation (128$\times$128$\times$48 TSDF volume) and the warped representation from last step $S'_{t-1}$ (the history aggregation part will includes more details). Twenty-two 3D convolution layers are applied to generate an output scene representation $S_{t}$ with a size of 8$\times$128$\times$128$\times$48.

\mypara{Action encoder (Fig. \ref{fig:network}b):}
In our setup, robots interact with the scene via pushing. The action can be discretized and represented by a tuple of integers $(p_x, p_y, d)$, where $p_x, p_y$ are the start coordinates of the push in Cartesian space, and $d$ represents one of $8$ pre-defined directions of a push. Inspired by prior work \cite{wu2020spatial}, we use the action map for input form in order to provide spatial alignment with scene representation. An action is represented as a one-hot matrix with a size of 8$\times$128$\times$128, where $[d, p_x, p_y]$ is $1$ and other places are filled with $0$. The action encoder encodes the action as two embeddings with size of 16$\times$32$\times$32 and 8$\times$64$\times$64.

\mypara{Motion prediction (Fig. \ref{fig:network}c):} 
% motion encoding (R,T)
The motion prediction network estimates transformations for every object based on the aggregated scene representation and action applied to the scene. The motion decoder predicts $k$ $\textbf{SE}(3)$ transforms, one for each of the predicted masks. We fix the last transformation ($k-1$) as an identical transformation since the background is static. A $\textbf{SE}(3)$ transform describes a rigid body transform $[R, t]$, specified by a rotation $R \in \textbf{SO}(3)$ and a translation $t \in \mathbb{R}^3$. Under this transformation, 3D point $x$ moves to $x' = Rx + t$. We represent rotations using a Euler transform vector.
Given the predicted $\textbf{SE}(3)$ transforms and masks, the transform layer produces a blended point cloud from the input points: { $y_j = \sum_{i=0}^{k-1} M_{ij} (R_i x_j + t_i)$}, where $y_j$ is the 3D output point corresponding to voxel $x_j$. Then the predicted scene flow of voxel $x_j$ is $y_j - x_j$. The motion loss $\mathcal{L}_\mathrm{motion}$ is Mean Square Error (MSE) between the predicted scene flow and ground truth. 

\mypara{Amodal instance mask prediction (Fig. \ref{fig:network}d):}
% output K channel, loss permute the ground truth
The mask predictor outputs a voxel-wise probability distribution $M_t$ over the $k$ classes. Following the standard practice as in prior works \cite{byravan2017se3,burgess2019monet}, we use $k=5$ as the maximum number of objects to be handled in our experiment. As shown in supplementary material, the trained model is able to generalize to test cases with fewer objects than the maximum number of objects.
During training, we need a specific order to calculate the loss for mask prediction and encourage temporal consistency over time.
At each step, we enumerate all the permutations and select the optimal matching and it serves as ground truth for the next step. Specifically, in the first step, its optimal matching is also used as ground truth for training right now.
Concretely, given the mask prediction and ground truth for each category, we calculate the negative log-likelihood loss: $\mathcal{W}_t(i, j) = M^{\mathrm{gt}}_t(i) \cdot \log M^{\mathrm{pred}}_t(j)$.
The loss of a matching is the summation of each category.
The optimal matching means it has the smallest matching loss: $\mathrm{match}_t = \argmin_{p} \sum_{i=0}^{k-1} \mathcal{W}_t(i, p(i))$
Once we find the correct order for the ground truth, the loss between predicted mask and ground truth $\mathcal{L}_{mask}$ is computed with Cross Entropy loss. 

% we compute a minimum-weighted bipartite graph matching between the mask prediction and ground-truth instances.
% Since the number of objects in our experiment is not very large, we enumerate the permutations $p_t$ of {$\{0, 1, \cdots, k-1\}$} to compute the optimal matching.
% Permutation enumeration can be replaced by the Hungarian algorithm to speed up the searching to handle a large number of objects. 
% The matching weight $\mathcal{W}_{i, j}$ is the negative log-likelihood between a pair of masks: 
% { $\mathcal{W}_t(i, j) = M^{\mathrm{gt}}_t(i) \cdot \log M^{\mathrm{pred}}_t(j)$}. The best order $p_t$ is the one with the smallest minimum-weighted matching score:
% { $\mathrm{match}_t = \argmin_{p} \sum_{i=0}^{k-1} \mathcal{W}_t(i, p(i))$}.  
% In the first step of each sequence, the $p_t$ is computed based on $\mathrm{match}_1$. For all the latter steps, $p_t$ is computed based on the matching of the previous step  $\mathrm{match}_{t-1}$ to encourage temporal consistency across steps. 
%\jw{I suggest we add the loss function.}

\mypara{Forward warping for spatially aligned history aggregation:}
To aggregate history, we warp the scene representation $S_t$ with the predicted scene flow  $F_t$ and mask prediction $M_t$ to produce features that are spatially aligned across multiple steps. Here we use trilinear interpolation for 3D warping because truncation or direct nearest neighbor wrapping results in more empty holes when several voxels are moving to the same position. Let {$(x^s_i, y^s_i, z^s_i)$} be the coordinates of voxel { $v^s_i$} in the input representation, {\small $(x^t_i, y^t_i, z^t_i)$} be the coordinates of a voxel { $v^t_i$} in the warped representation, and {\small $(x^f_i, y^f_i, z^f_i)$} is the predicted scene flow of { $v^s_i$}. The weight contribution of { $v^s_i$} to { $v^t_j$} is computed by:
{\small
$$
    W_{v^s_i \rightarrow v^t_j} = m_i \cdot \max(0, 1 - |x^s_i + x^f_i - x^t_j|) \cdot \max(0, 1 - |y^s_i + y^f_i - y^t_j|) \cdot \max(0, 1 - |z^s_i + z^f_i - z^t_j|),
$$}
where {\small $m_i=\sum_{d=0}^{k-2} M_t[d, x^s_i, y^s_i, z^s_i]$}  is the predicted probability that $v^s_i$ belongs to any object. The last channel (d=k-1) always represents empty space. Let {\small $S_t(i)$} represent the input feature value at $v^i$,
then output the feature {\small$S'_{t}(j)$} at {\small $v^t_j$}  after warping is computed as:
{\small $ S'_{t}(j) = (\sum_i S_t(i) \cdot W_{v^s_i \rightarrow v^t_j}) / \sum_i W_{v^s_i \rightarrow v^t_j}$}.
\mypara{Loss function.} The final loss function is $\mathcal{L} = \mathcal{L}_\mathrm{motion} + \alpha\mathcal{L}_\mathrm{mask}$, where $\mathcal{L}_\mathrm{motion}$ is the Error of motion prediction and $\mathcal{L}_\mathrm{mask}$ is the loss of mask prediction and $\alpha=5$ is a weighting factor.

\subsection{Applying DSR in Robot Manipulation}
We now demonstrate how DSR can be used in manipulation. Specifically, the goal of the task is to control a robot arm to push objects in the scene to match a target state. With our learned model, we perform temporally extended planning by choosing a sequence of actions that can be executed in the environment. Among different planning approaches, we choose model-predictive control (MPC) to take advantage of our predictive model. 

We apply a simple shooting-based MPC method \cite{nagabandi2017modelbased} to generate and plan for a sequence of actions that minimize the cost. First, we sample actions around predicted masks from our DSR model, since only these actions are close to the objects. This allows us to have a much smaller sample size of actions and make our decision making faster. Then, we compute the cost based on the next state predictions, which include the pivot points and masks in the next state. Specifically, we have {\small $ \mathrm{cost}(a_1, a_2, ..., a_n) = \sum_{i} (\lambda_{i} \times L^\mathrm{pos}_i - \mathrm{IoU}_i) $}, where $a_1, a_2, ..., a_n$ are candidate actions, { $L^\mathrm{pos}$} is the Mean Square Error between target and predicted positions (computed by predicted mask) of each object, { $\mathrm{IoU}_i$} is the IoU between the predicted mask and target state, $\lambda_i$ is a weighting factor for each channel.
Finally, we choose the sequence of actions that has the lowest cost. 

\begin{wrapfigure}{r}{0.6\textwidth}
\vspace{-5mm}
  \begin{center}
  \includegraphics[width=0.6\textwidth]{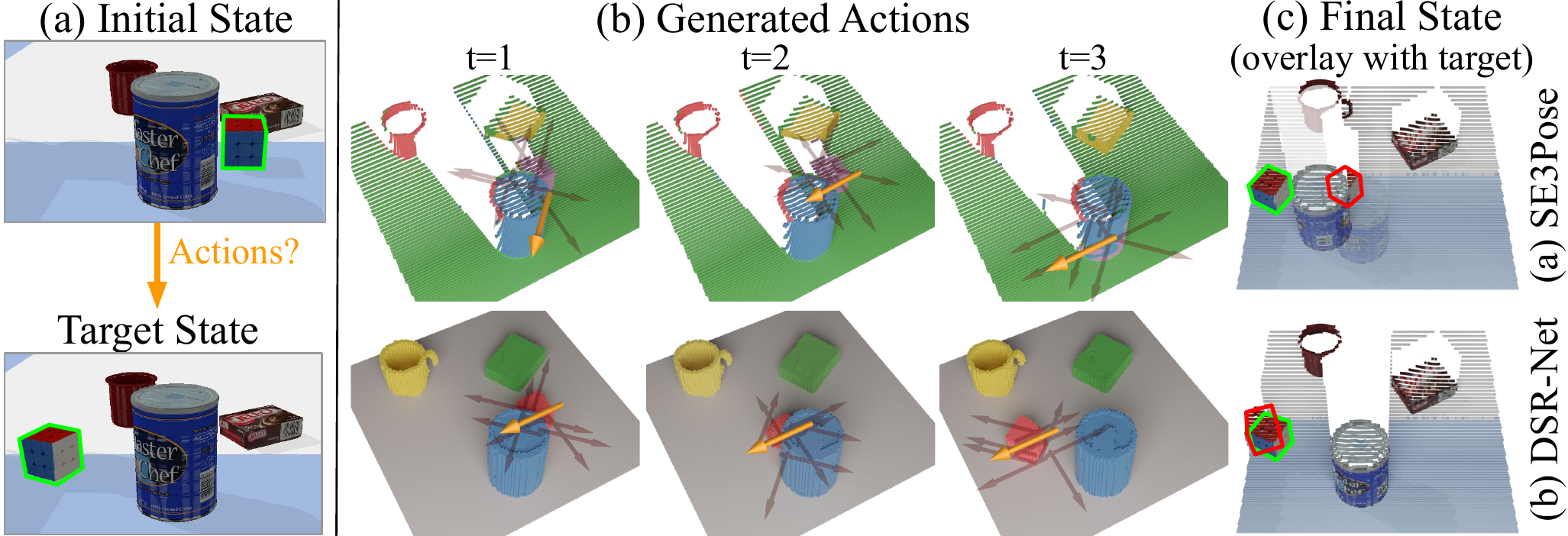}
  \end{center}
  \vspace{-3mm}
  \caption{\textbf{Application of DSR in Planner Pushing.} (a) The goal is to generate a sequence of actions to push objects to match a target state. (b) In each step, a set of action candidates are sampled and the action with the lowest cost (yellow) is chosen to execute. At $t=3$, SE3Pose-Net loses track of the occluded object hence choose the wrong action, while DSR-Net correctly models the occluded object and chooses appropriate actions. (c) The final state (red) of DSR-Net is much closer to the target state (green).}
  %https://docs.google.com/drawings/d/1KB66jVLI1F6zCgWjVcqPBEgT3ocOKi_d2cIxnSlzcYA/edit
  \vspace{-3mm}
  \label{fig:mpc}
\end{wrapfigure}

Since DSR maintains object permanence in the representation, it enables the planning algorithm to use the full state information including the occluded objects. 
For example, in Fig. \ref{fig:mpc}, the robot has to push an occluded cube to a target location. This is impossible with SE3-Pose-Net -- since it models only visible surfaces, the object is completely missing due to occlusion (t=3) and therefore a wrong action is selected. With DSR, the control policy is able to sample actions around the occluded object to predict the next state and cost accurately. Quantitative result are shown in Sec. \ref{sec:eval_mpc}.

\vspace{-2mm}
\section{Dynamic Scene Representation (DSR) Benchmark} \label{sec:data}
\vspace{-2mm}
To quantitatively evaluate predictive models, we need a dataset that contains robot interactions and ground object motion. 
Since there is no existing dataset containing this information (especially with real-world robot interactions), we construct a new dynamic scene representation (DSR) benchmark that contains both simulation and real-world data for training and evaluation. 
%Since there are not existing dataset contains real-world robot interaction and ground object motion, we construct a dynamic scene representation (DSR) benchmark. 

\mypara{Simulation data.}
We use two types of objects in simulation training data: (1) cubes with different sizes $s \in [0.02, 0.04]$m, and (2) 44 shapenet objects of 5 categories: mug (5), bottle (14), can (6), phone (10), and sofa (9).
For each sequence, we choose 4 objects and randomly drop them on the workspace ($0.512$m $\times$ $0.512$m). Then the robot executes 10 random pushing actions with a simple heuristic-base policy that encourages the change of spatial order and prevents moving objects out of the workspace. Details of the policy are described in the supplementary material.
In total, there are 8,000 sequences with 80,000 interactions for training.
We also generate a testing dataset using YCB objects~\cite{ye2019object} with the same interaction policy. This includes 400 sequences with 4,000 interactions.

\begin{wrapfigure}{r}{0.6\textwidth}
\vspace{-6mm}
  \begin{center}
   \includegraphics[width=0.6\textwidth]{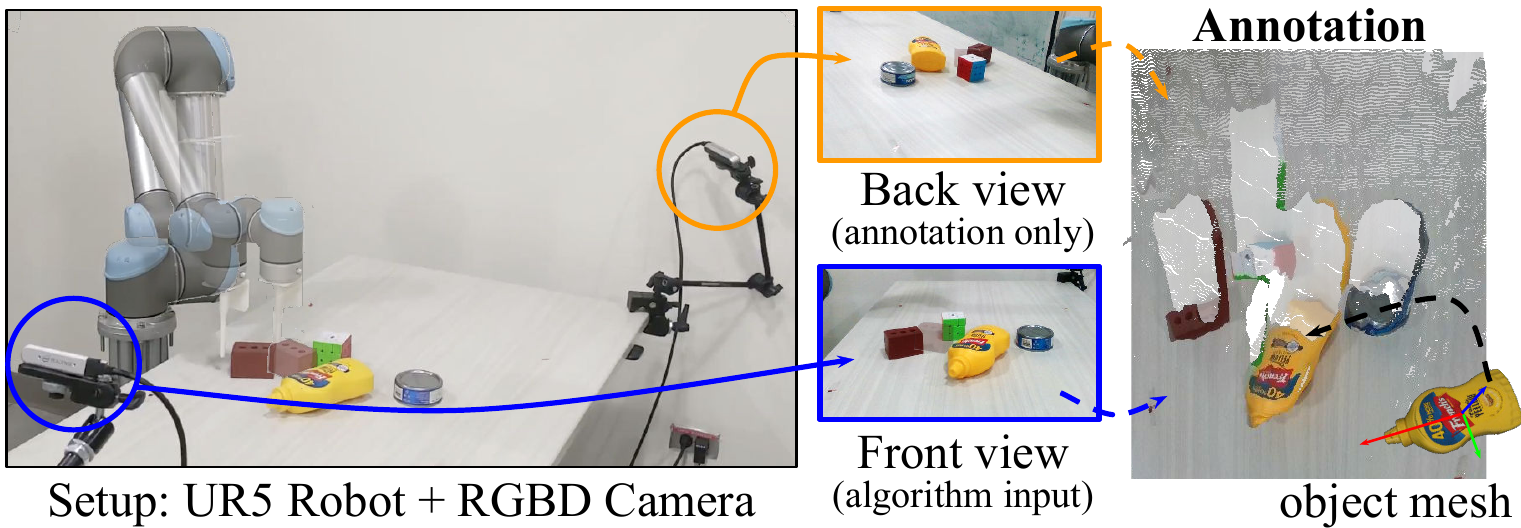}
  \end{center}
  \vspace{-3mm}
  \caption{\textbf{Realworld Setup and Annotation UI.} \textbf{[left]} UR5 is used for robot manipulation. \textbf{[middle]}  We capture RGB-D images using two calibrated Intel RealSense D415. The front view image is taken as input by the algorithm and the back view image is only used for annotation. \textbf{[right]} The object mesh is moved with keyboard to match the fusion of point clouds from two cameras.}
  %https://docs.google.com/drawings/d/19U33yPFtVaM8yjXfJOn4VpsXdjXTuEQHavWuPkpBltM/edit
  \vspace{-3mm}
  \label{fig:real_setup}
\end{wrapfigure}

\mypara{Real-world data.}
\label{sec:exp_mpc}
Our real-world setup consists of a UR5 robot with a cylinder pusher tool and two calibrated RGB-D cameras. Fig. \ref{fig:real_setup} shows the setup and YCB objects \cite{ye2019object} used in the real-world experiments. 
To accurately annotate the ground truth object pose under occlusion, we use an additional calibrated RGB-D camera in the setup to provide a backview of the workspace (Fig. \ref{fig:real_setup} left and middle). We use the same discrete action space to collect real data.
During annotation, we combine the 3D point clouds from both views to obtain a complete observation for the entire workspace. 
Fig. \ref{fig:real_setup} right shows our annotation user interface. Users can control the object meshes' 6DoF poses with keyboard to match the pose in the scene.  In total, we collect 90 sequences with 900 interaction steps. 

\mypara{Train/Test split.}
In simulation, all models are trained with 8,000 sequences with ShapeNet objects and tested on 400 sequences with novel YCB objects.  
The real-world dataset is split into 50 finetuning sequences and 100 testing sequences. In the following experiments, models labeled with ``ft'' are finetuned with the 50 sequences, all the other models are directly tested with realworld data without finetune. All qualitative result (except SE3 and SE3PoseNet) are using model without finetune. 
% all models are further finetuned with 30 real-word sequences and then tested with another 60 real-world sequences. 

\vspace{-2mm}
\section{Evaluation}
\vspace{-2mm}
\label{sec:eval}
We designed a series of experiments in both simulation and the real world using the benchmark data described in Sec. \ref{sec:data} to validate design decisions, and to compare with other models that predict future scene representations. Specifically, we want to see whether \OURS is able to 
\begin{itemize}
    \vspace{-2mm}
    \item[]  \hspace{-9mm} (1) Accurately predict object motion under different robot interactions;
    \item[]   \vspace{-1mm} \hspace{-9mm} (2) Aggregate the history and encodes object   \textit{permanence} and \textit{continuity}; and
    \item[]   \vspace{-1mm} \hspace{-9mm} (3) Improve the performance of down-stream manipulation tasks.
\end{itemize}

\begin{wraptable}{r}{0.48\linewidth}
\vspace{-9mm}
{\small
\setlength\tabcolsep{1.5pt}
    \begin{tabular}{clcc|cc}
        \toprule
            & & \multicolumn{2}{c|}{Simulation} & \multicolumn{2}{c}{Real}\\
            & & visible & full & visible & full \\
            \midrule
            \multirow{3}{*}{2D}  & \BaselinePixelFlow ft \cite{byravan2017se3}       & 8.24 & - & 7.63 & - \\
                                    & \BaselineSE ft \cite{byravan2017se3}      & 7.84 & - & 6.91 & - \\
                                    & \BaselineSEPose ft \cite{byravan2018se3}  & 13.01 & - & 10.49 & - \\
            \midrule
            \multirow{4}{*}{3D}  & \BaselineVoxelFlow   & 7.34 & 0.093 & 6.80 & 0.094\\
                                    & \BaselineSingle   & 5.94 & 0.086 & 6.64 & 0.093 \\
                                    & \OURS       & \textbf{5.54} & \textbf{0.082} & {6.51} & {0.090 }\\
                                    & \OURS ft      & - & - & \textbf{3.33} & \textbf{0.048}\\
        \bottomrule
    \end{tabular}
    \vspace{-1mm}
    \caption{\textbf{Average flow Error (MSE in cm)}}
    \label{tab:motion}
    \vspace{-5mm}
    }
\end{wraptable}

\subsection{Motion Prediction}\vspace{-2mm}
First we want to evaluate the learned scene representation on predicting object motion under robots' interaction. 
We use the Mean Squared Error (MSE) to evaluate the predicted 3D scene flow. 
For image based approaches, the MSE is computed and averaged for pixels of the visible surface, same as in SE3-Net~\cite{byravan2017se3}. 
For voxel-based approaches, the MSE is computed and averaged over the voxels on visible surfaces of the object (visible) and all voxels (full) separately. %\jw{I don't understand.}

\begin{figure*}[t]
\centering
\includegraphics[width=\linewidth]{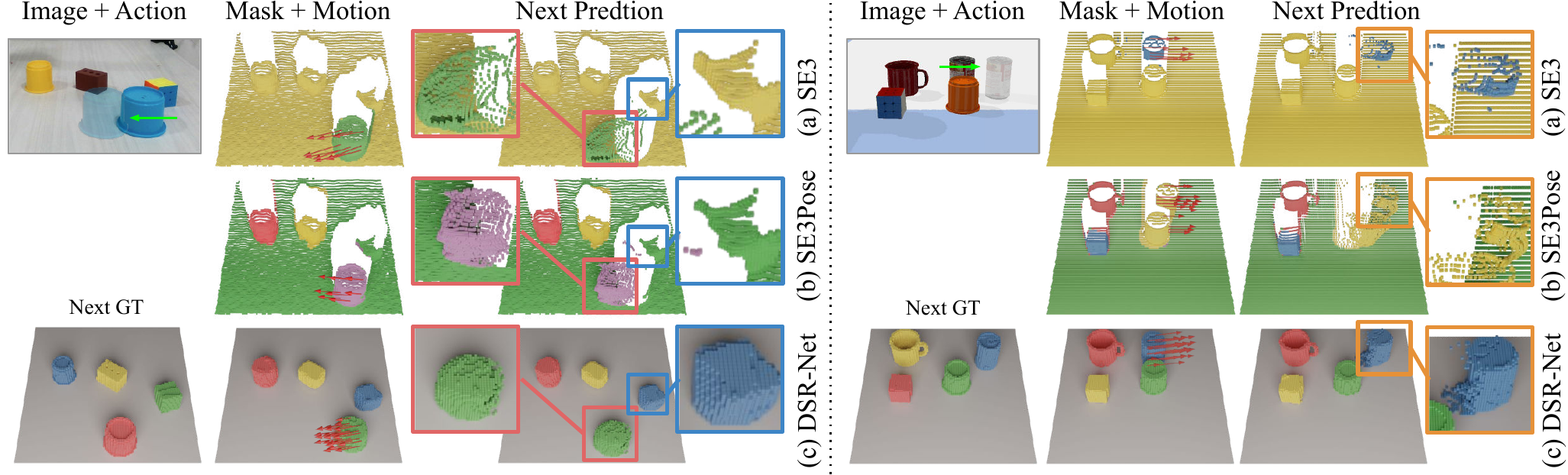}
\caption{\textbf{Amodal Mask and Motion Prediction.} The mask and motion prediction of \BaselineSE and \OURS in both real-world (left) and simulation (right). \BaselineSE predicts only masks for moving objects and the estimated motion is limited to the visible surface. Although the mask prediction in \BaselineSEPose is not limited to moving objects, it fails to separate closed objects and miss small objects. \OURS produces the full 3D volume as well as masks for all objects in the scene.}
\label{fig:result}
\vspace{-5mm}
\end{figure*}

\mypara{Baselines:} In this experiment, we compare our algorithm with the following predictive models: 
\begin{itemize}
\vspace{-3mm}
\item[\textbullet ] \BaselinePixelFlow~\cite{byravan2017se3}: it predicts per-pixel scene flow for the visible surface. 
\vspace{-1mm}
\item[\textbullet] \BaselineSE~\cite{byravan2017se3}: it predicts per-object masks and SE3 motions
\vspace{-1mm}
\item[\textbullet] \BaselineSEPose~\cite{byravan2018se3}: it predicts per-object poses, masks, and SE3 motions
\vspace{-1mm}
\item[\textbullet] \BaselineVoxelFlow: it predicts per-voxel scene flow for the entire 3D volume. 
\vspace{-1mm}
\item[\textbullet] \BaselineSingle:  \OURS without history aggregation. 
%\item[\textbullet] \OURS: Our model shown in Figure \ref{fig:network}.
\vspace{-1mm}
\end{itemize}

\mypara{Compared with state-of-the-art predictive models.}
Tab. \ref{tab:motion} shows quantitative comparisons of predicted motion.
Since most voxels are static, the error of full volume is much smaller than the visible surface error.
Fig. \ref{fig:result} shows qualitative comparisons among our model, \BaselineSE, and \BaselineSEPose. The visualization suggests that the motion estimation in \BaselineSE is limited to visible surface and cannot model occluded regions.
In addition, the mask prediction in \BaselineSE only handles the moving object, treating all other objects as background. 
This is because \BaselineSE predicts masks based on both observation and action, where the network learns to first identify the moving objects and then predict mask and motion for these objects only. 
The mask prediction of \BaselineSEPose is independent from action; therefore, it has to predict masks for all objects in the scene. However, the motion prediction of \BaselineSEPose is based on object poses without considering their detailed geometry. This fact makes \BaselineSEPose perform worse in motion prediction.
In contrast, our model produces 3D amodal masks for all objects in the scene and predicts the object motion  more accurately.

% \mypara{Effect of the motion representation.} 
% Our network predicts SE3 transformation for each individual object, which encourages rigidity within one object instances. Comparing to \BaselineVoxelFlow  that does not not have this constraint, we can observe that by representing the scene as individual objects and predicting SE3 transformation for each of them results in a much accurate motion prediction. The result demonstrates that enforcing rigidity in the motion prediction process is effective way of improving the prediction quality. \jw{I'm not sure if you'd like to make this an argument. This also limits your model's capacity in dealing with deformable objects.}

% \mypara{Effect of history aggregation in motion prediction.} 
% Compared with \BaselineSingle, our model has better motion prediction performance by aggregating history information (Tab. \ref{tab:motion}). This advantage owes to the ability to model occluded objects with history aggregation, as shown in Fig. \ref{fig:consistent} and \ref{fig:persistent}. As our real-world dataset only contains a few cases where an object moves into an occluded region, the quantitative improvement on real data is small. 
% The improvement on simulation data is larger, because we design the environment to focus on these corner cases. %\jw{I'd like to see numbers. This reads speculative.} 

\vspace{-2mm}
\subsection{Temporal Information Aggregation}\vspace{-2mm}

% \begin{wraptable}{r}{0.47\linewidth}
% {\small
% \centering
% \vspace{-5mm}
% \setlength\tabcolsep{1.5pt}
%     \begin{tabular}{lcccc}
%         \toprule
%             & \multicolumn{2}{c}{Simulation} & \multicolumn{2}{c}{Real}\\
%             & unordered & ordered & unordered & ordered \\
%             \midrule
%             GTWarp        & 0.802 & 0.802 & 0.674 & 0.673 \\
%             \midrule
%             \BaselineSingle   & 0.720 & 0.480 & 0.644 & 0.507\\
%             \BaselineNoWarp        & 0.724 & 0.710 & 0.635 & 0.626\\
            
%              \OURS &\textbf{ 0.736} & \textbf{0.726} & \textbf{0.647} & \textbf{0.642}\\
%         \bottomrule
%     \end{tabular}
%     \vspace{-2mm}
%     \caption{\textbf{Amodal Object Mask Prediction IoU}}
%     \label{tab:mask}
% \vspace{-3mm}
% }
% \end{wraptable}

\begin{wraptable}{r}{0.47\linewidth}
{\small
\centering
\vspace{-7mm}
\setlength\tabcolsep{1.5pt}
    \begin{tabular}{lcccc}
        \toprule
            & \multicolumn{2}{c}{Simulation} & \multicolumn{2}{c}{Real}\\
            & unordered & ordered & unordered & ordered \\
            \midrule
            GTWarp        & 0.807 & 0.807 & 0.646 & 0.645 \\
            \midrule
            \BaselineSingle & 0.753 & 0.526 & 0.613 & 0.485\\
            \BaselineNoWarp & 0.762 & 0.756 & 0.625 & 0.624\\
            
             \OURS &\textbf{ 0.772} & \textbf{0.767} & \textbf{0.628} & \textbf{0.628}\\
        \bottomrule
    \end{tabular}
    \vspace{-2mm}
    \caption{\textbf{Amodal Object Mask Prediction IoU}}
    \label{tab:mask}
\vspace{-3mm}
}
\end{wraptable}

In this section, we evaluate whether \OURS is able to effectively aggregate the history information to capture object permanence and continuity. 
We use two types of intersection over union (IoU) scores on 3D amodal instance masks as the evaluation metric: unordered and ordered IoU.  
To compute unordered IoU, we obtain the object order for each step by permuting the objects' order and use the one that maximizes the average IoU over all objects as the ground truth order. 
The order of step $t$ is calculated by {\small {$\mathrm{order}_t = \argmax_{p} \frac{1}{k}\sum_{i=0}^{k-1} IoU_t[M^\mathrm{gt}_t(i), M^\mathrm{pred}_t(p(i))]$}}, where $k$ is the number of objects.
For ordered IoU, we permute the object instance index once and use the order that best matches the entire sequence as the ground truth order:
{\small {$\mathrm{order}_t = \argmax_{p} \sum_{s=0}^{N-1} \frac{1}{k}\sum_{i=0}^{k-1} IoU_s[M^\mathrm{gt}_t(i), M^\mathrm{pred}_s(p(i))]$}}, where $N$ is the number of interactions.
To achieve a high ordered IoU, the system needs to maintain a consistent order of object instance throughout the interaction steps. Therefore, this metric reflects the continuity of the scene representation over time. 
Besides, since the 3D IoU is evaluated on all the voxels in the scene regardless of occlusion, this metric also naturally measures the permanence of the scene representation under occlusion.

\begin{figure*}[t]
\centering
\includegraphics[width=\linewidth]{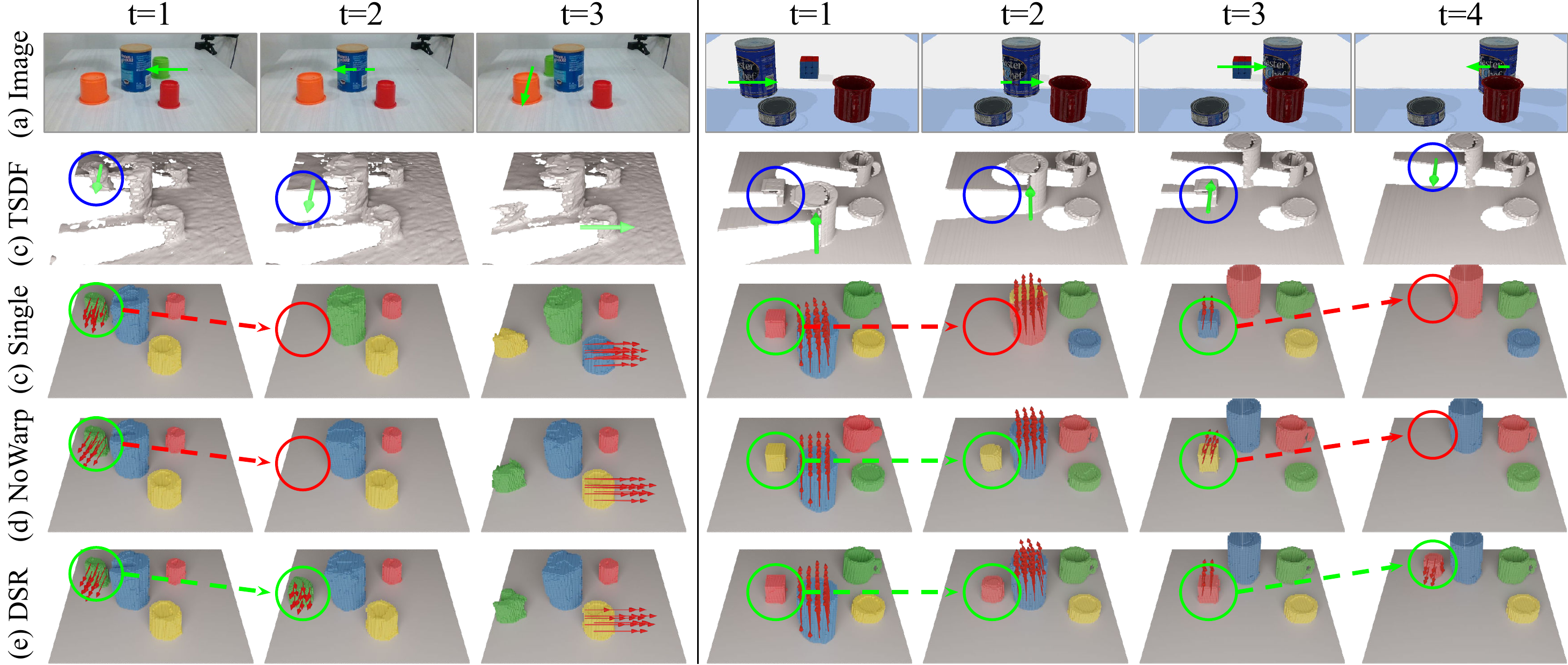}
\caption{\textbf{Scene Representation with Object Permanence.} The TSDF and result are rendered in side view to better show occlusion cases. Object permanence is labeled in green circle and failure cases are labeled in red. 
At t=2, in the real-world example (left), the green cup is occluded by the can. Only \OURS is able to predict the permanence of the green cup. In the simulation example (right), occlusion appears in the t=2 and t=4. The difference is that at t=1-2, the Rubik's Cube is static when being occluded, and at  t=3-4, the Rubik's Cube is moved and then occluded (dynamic occlusion). 
\BaselineSingle fails in both cases.  \BaselineNoWarp  can handle the first case since the history contains the information of the static Rubik's Cube, but cannot handle dynamic occlusion due to the lack of motion in the history.  \OURS is able to handle both  cases.}
\label{fig:persistent}
\vspace{-5mm}
\end{figure*}

\mypara{Baselines.}
In this experiment,we compare our aggregation model with following alternatives: 
\begin{itemize}
\vspace{-3mm}
    \item[\textbullet] \BaselineSingle: it does not use any history aggregation.
    \vspace{-1mm}
    \item[\textbullet] \BaselineNoWarp: it does not warp the representation before aggregation.
    \vspace{-1mm}
    \item[\textbullet] \BaselineGTWarp: it warps the representation with ground truth motion (i.e., performance oracle). 
    \vspace{-1mm}
\end{itemize}

\mypara{Does history aggregation help in on amodal shape completion?} 
The unordered IoU in Tab. \ref{tab:mask} measures the quality of 3D amodal shape completion without consider the objects' identity. 
The result demonstrates that by effectively aggregate the past observations, our method is able to infer a more accurate scene representation in terms of modeling an object's complete 3D geometry  from partial observations (+1.6\% improvement in unordered IoU compare to the single step model). 
In the following experiments, we will evaluate the object permanence and continuity using ``ordered IoU''. 

\mypara{Does DSR encode object permanence?}
To evaluate the object permanence, we examine the network's ability to infer an object's existence during occlusion.  
Fig. \ref{fig:persistent} shows amodal mask estimation under occlusion in both real world and simulation cases.
There are two static-occlusion cases (t=1-2 in the real and simulation example), where the occluded object is not moving, and one dynamic-occlusion case (t=3-4 in the simulation), where the moving object becomes occluded.
The \BaselineSingle model fails in both. The \BaselineNoWarp model can handle one of static occlusion cases, since the history contains the information of the static object. However, it cannot handle dynamic occlusion due to the lack of historical motion. \OURS is able to handle both static and dynamic occlusions.

\begin{figure*}[t]
\centering
\includegraphics[width=\linewidth]{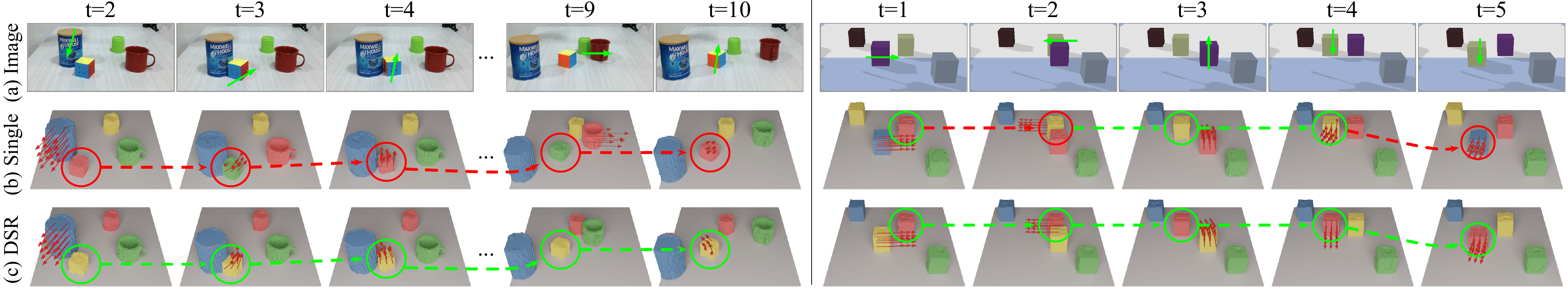}
%https://docs.google.com/drawings/d/1kWvv1O3mL6bk5QnVXyMo3hqixJhmbFVWTX_HGMT3N9g/edit?usp=sharing
\caption{\textbf{Scene Representation with Object Continuity.} The mask prediction of \BaselineSingle model (b) and DSR-Net (c) after several interactions. Continuous instance prediction between two consecutive steps are highlighted in green, while discontinuity is highlighted in red. In the simulation example, four identical cubes are  indistinguishable in depth and the two cubes swap their positions during interaction. \OURS can track objects even when the spatial order is significant changed during interactions, while the \BaselineSingle model fails.}
\label{fig:consistent}
\vspace{-5mm}
\end{figure*}

\mypara{Does DSR encode object continuity?}
% what is consistency, why it is important 
A model that captures spatiotemporal continuity should maintain a consistent object identity overtime. We evaluate this and show the results in the ordered IoU in Tab. \ref{tab:mask}. 
Fig. \ref{fig:consistent} presents qualitative results of mask prediction after several interactions. Unlike the \BaselineSingle model, which is sensitive to the spatial order, our model maintains spatiotemporal continuity via consistent labeling of object instances. In the simulation demonstration, \OURS can even track visually indistinguishable objects, whose positions are swapped after several interactions. It proves that the continuity owes to history aggregation, instead of visual appearance. Note that this happens to align with classical findings in developmental psychology~\cite{spelke2007core}.
The small gap between unordered and ordered IoUs in Tab. \ref{tab:mask} demonstrates that our \OURS achieves a consistent order of object instances during an interaction sequence; \BaselineSingle has a much bigger gap, indicating that it fails to track object identity without history aggregation. %\jw{why does this matter?}

\mypara{Does motion prediction help in history aggregation?}
To test the effect of motion prediction on history aggregation, we compare our model with \BaselineNoWarp. 
The plot in Tab. \ref{tab:mask} shows that, with spatial-aligned features, the algorithm produces a more accurate scene representation. In both simulation and real-world test sets, \OURS consistently achieves higher order and unordered IoUs.
Further, if we warp the scene representation with ground truth motion as in \BaselineGTWarp, the algorithm achieves even higher performance. 
Thus, warping features with correct object motion is helpful for aggregating history information. We conjecture that this is because the warping operation provides a spatially aligned feature representation of current and next states, making the information aggregation easier. 

\vspace{-2mm}
\subsection{Apply DSR in Robot Manipulation \label{sec:eval_mpc}} \vspace{-2mm}
Finally, we evaluate the performance of using DSR in planer pushing, where the goal is to generate an action plan of a robot arm to push objects in the scene to match a target configuration. We compare the performance of our DSR model with  \BaselineSE~\cite{byravan2017se3} and \BaselineSEPose~\cite{byravan2018se3} using 100 target states collected from the simulation environment. We used the planning method described in Sec. \ref{sec:exp_mpc} to generate action sequences with a length of 3 to match a pre-collected target state. Then, we compute the voxel IoU between the final full states and the groundtruth target states for evaluation.
In this task, our model achieves a \textbf{0.72} IoU, outperforming \BaselineSE and \
\BaselineSEPose, whose IoU are \textbf{0.31} and \textbf{0.32} respectively. Thus, using \OURS with MPC results in better state matching with target.

%While \BaselineSEPose produces reasonable masks from partial observations (Fig. \ref{fig:motion}), the positions predicted from these masks are not accurate, since they tend to bias towards the visible parts of the masks (Fig. \ref{fig:mpc}). Producing a full mask of each object, our model can estimate positions more accurately. Further, as shown in Tab. \ref{tab:motion}, \OURS does better in motion prediction. Combining better masks and motion predictions, using \OURS with MPC results in better state matching with target configurations. %\jw{Any results support the last sentence?}

\vspace{-2mm}
\section{Conclusions}
\vspace{-2mm}
We have introduced a new 3D dynamic scene representation that, by design, captures object permanence, solidity, and spatio-temporal continuity.
We have also proposed \OURS, an end-to-end framework that learns to aggregate information over multiple interactions to build such a representation from visual observations.
%To train and evaluate the proposed algorithm, we have presented a new benchmark with over 80,000 simulated interactions and 900 real-world interactions. 
Our experiments in both simulation and real world show that \OURS achieves state-of-the-art performance in modeling 3D scene dynamics and enables more accurate action planning in an object pushing task. 

%===============================================================================

% The maximum paper length is 8 pages excluding references and acknowledgements, and 10 pages including references and acknowledgements

\clearpage
% The acknowledgments are automatically included only in the final version of the paper.
\acknowledgments{We would like to thank Google for providing the UR5 robot hardware for our experiments. This work was supported in part by the Amazon Research Award, the Columbia School of Engineering, as well as the National Science Foundation under CMMI-2037101.}

%===============================================================================

% \bibliography{example}  % .bib
% \bibliographystyle{plainnat}
{\small
\bibliography{egbib}
}

\newpage
\appendix
\renewcommand{\thesection}{A.\arabic{section}}
\renewcommand{\thefigure}{A\arabic{figure}}
\renewcommand{\thetable}{A\arabic{table}}
\setcounter{section}{0}
\setcounter{figure}{0}
\setcounter{table}{0}

\section{Interaction Policy}
We use a heuristic-base policy to encourages the change of spatial order and prevent moving objects out of the workspace. The interaction policy includes two steps: choose an object and choose the direction. 

Each object has a score that is initialized as 0. The softmax of the score is considered as the probability to be chosen in each step. In each step, the value of the chosen object increased by one, and if the value is larger than 2, it becomes -2. Locally, this strategy leads to an object being pushed twice consecutively, which encourages to the change of spatial order. Globally, it maintains a balance among all the objects. 

After choosing the object, each direction is also assigned a score: $Q(\vec v) = 1.5 (\overrightarrow v \cdot \overrightarrow{p_0, p_t}) + 2 (\overrightarrow v \cdot \overrightarrow{p_{t-1}, p_t})$, where $p_0, p_{t-1}, p_t$ are the initial, last and current position respectively. This can encourage the object to move away from its initial position and prevent from being pushed back and forth. Also, if the object is object is far away from the workspace center (distance $\geq 0.2$ ), those directions making it further 
away will get a punishment of $-10$ to prevent the object out of workspace. Again, the probabilities of each direction are the softmax of their score values.

\section{Network Structure}
The scene encoder concatenates the current observation ($128\times128\times48$ TSDF volume) and the warped representation from last step $S'_{t-1}$ with a size a $8 \times128\times128\times48$. It first applies ten $3\times3\times3$ convolution layers with $16$, $32$, $32$, $32$, $32$, $64$, $64$, $64$, $64$, and $128$ channels. The strides sizes of the first, second, sixth and tenth layer are $2\times2$. Between convolution layers, there are batch normalizations, Leaky ReLUs with slope 0.2. The outputs of the first, fifth and ninth layer are also reserved for skip connection. 
After two residual blocks, eight $3\times3\times3$ convolution layers are applied with $64$, $64$, $32$, $32$, $16$, $16$, $8$, and $8$ channels. The input of the third, fifth, and seventh are concatenated with reserved tensors for skip connections. The output of the first, third, fifth, and ninth layer are upsampled by $2\times$ with trilinear upsampling. Finally, the scene representation $S_t$ will be $8\times128\times128\times48$.

The action encoder takes an action map as input, with a size of $8\times128\times128$. Nine $3\times3$ convolution layers are applied with $64$, $64$, $64$, $64$, $128$, $128$, $128$, $128$, and $16$ channels. The first and fifth layer's stride sizes are $2\times2$. Another $3\times3$ convolution layer with 8 channels are applied after the fourth layer to generate an embedding with different size. Thus, the action map is encoded as two embeddings with size of $16\times32\times32$ and $8\times84\times64$.

The mask predictor consists of a $1\times1\times1$ convolution layer and a softmax layer to outputs a per-voxel mask probability distribution.

The motion predictor takes the scene representation $S_t$ and action embedding as input. It first applies eight $3\times3\times3$ convolution layers with $8$, $16$, $32$, $32$, $32$, $64$, $128$, and $128$ channels, five of which have strides sizes $2\times2$. The input of the first three layers are also concatenated with original action and two action embeddings respectively. The 2D tensor is repeated in the last channel to concatenate with 3D tensor. Then a 3D convolution layer with kernel size $4\times4\times2$ is applied to the $128\times4\times4\times2$ feature to generate a vector with a length of $128$. After five fully connected layers with $512$ hidden units, $k\textbf{SE}(3)$ transforms are output, one for each predicted mask.

\section{Training Details.}
We implement our model in PyTorch.
Optimization is carried out using ADAM with $\beta_1$ = 0.9 and $\beta_2$ = 0.95. The model is trained with a minibatch of 48 for 30 epochs in the first stage and 20 epochs for the other two stages. The whole training takes about 20h on 4 NVIDIA 2080Ti GPUs. The inference speed is around 15 fps on a single GPU.

Since the aggregation ability depends on the accuracy of motion prediction, we split the training process into three stages from easy to hard: (1) single-step on cube dataset; (2) multi-step on cube dataset; (3) multi-step on ShapeNet dataset. 
In the first stage, we use an initial learning rate of $10^{-3}$ and a learning rate decay of $0.5$ after each $5$ steps. In the other two stages, we finetune the previous model with an initial learning rate of $10^{-4}$ and a learning rate decay of $0.5$ after each $3$ steps.

We also include a table of symbols that is used in our algorithm section:
\begin{table}[H]
{
\centering
    \begin{tabularx}{.8\textwidth}{lX}
        \toprule
            $S_t$ & volumetric scene representation at step $t$. \\
            $S’_t$  & volumetric scene representation warped with scene flow prediction at step $t$.\\
            $M_t / M^{gt}_t$ & volumetric amodal instance mask prediction / ground truth at step $t$.\\
            $F_t / F^{gt}_t$ &volumetric scene flow prediction / ground truth at step $t$\\
            $match_t$ & optimal matching between mask prediction as ground truth at step $t$ \\
            $k$ & maximum number of objects (including background) \\
            $(p_x, p_y, d)$ & input action, where $p_x$ and $p_y$ are the start coordinate of the push, and $d$ is the direction index. \\
            $[R, t]$& SE(3) transformation, where R is a rotation matrix, and $t$ is a translation vector. \\
            $W_{i \rightarrow j}$& weight contribution of voxel $i$ to voxel j in forwarding warping.\\
            $L_{motion}, L_{mask}$ & motion loss and mask loss used for training.\\
            $L_{pos}$ & position error between target and prediction used for planner pushing.\\
        \bottomrule
    \end{tabularx}
    \caption{A table of symbol used in the paper}
}
\end{table}

\section{Action Sampling for MPC}
We integrate our model with a shooting-based MPC approach for planner pushing. 
We firstly sample a batch of actions for each for cost calculation. Since \OURS and \BaselineSEPose can produce a mask of each object in the scene, we only sample around the masks to reduce sampling size. For \BaselineSE, we perform uniformly random sampling under the whole action space. Specifically, we sample 100 actions for \OURS and \BaselineSEPose and 200 actions for \BaselineSE. We then use this action batch and current observation as input to retrieve outputs from the models mentioned above. Lastly, we calculate the cost of each of the output using the cost function specified in Sec. 3.1 and choose a sequence of actions that has the smallest cost. While we calculate costs with all objects for \OURS and \BaselineSEPose, we only calculate the cost between the mask from \BaselineSE, and the moving object for \BaselineSE only produces masks for moving objects.

\section{Dataset Collection}
We collected our benchmark data set on a similar real setup as our simulation environment. An action is picked by an in-house human expert to create sequences of object scenes. Specifically, we used the following list of object in YCB dataset: sugar box, tomato soup can, mug, chocolate pudding box, gelatin box, potted meat can, chips can, coffee can, cracker box, bleach cleanser, enamel-coated metal bowl, spring clamps, plastic banana, plastic orange, foam brick, different sizes of cups, Lego Duplo, Rubik’s cube.

\begin{figure}[t]
    \includegraphics[width=\linewidth]{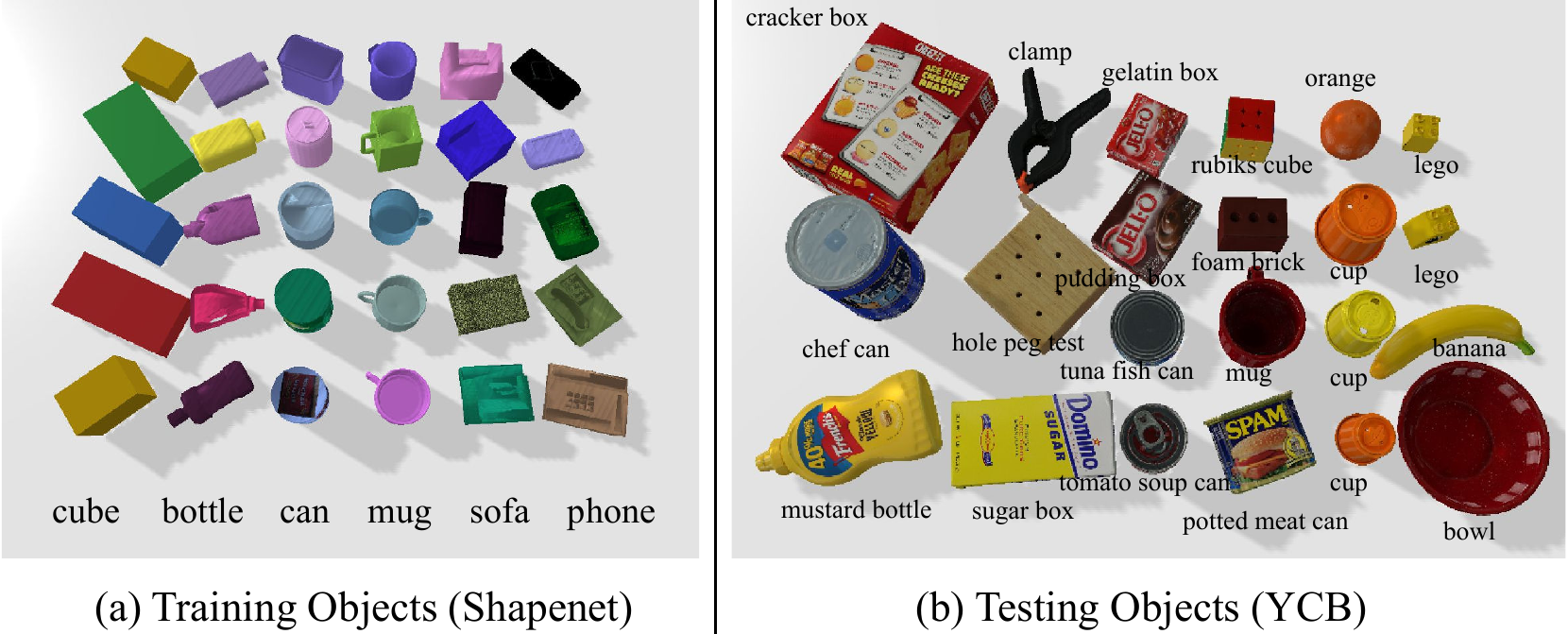}
    \caption{ \textbf{Objects.} (a) training objects from Shapenet dataset. (b) testing objects from YCB dataset.}
    \label{fig:object}
\end{figure}

\section{Generalization to Different Number of Object}
\begin{figure}[t]
    \centering
    \includegraphics[width=\linewidth]{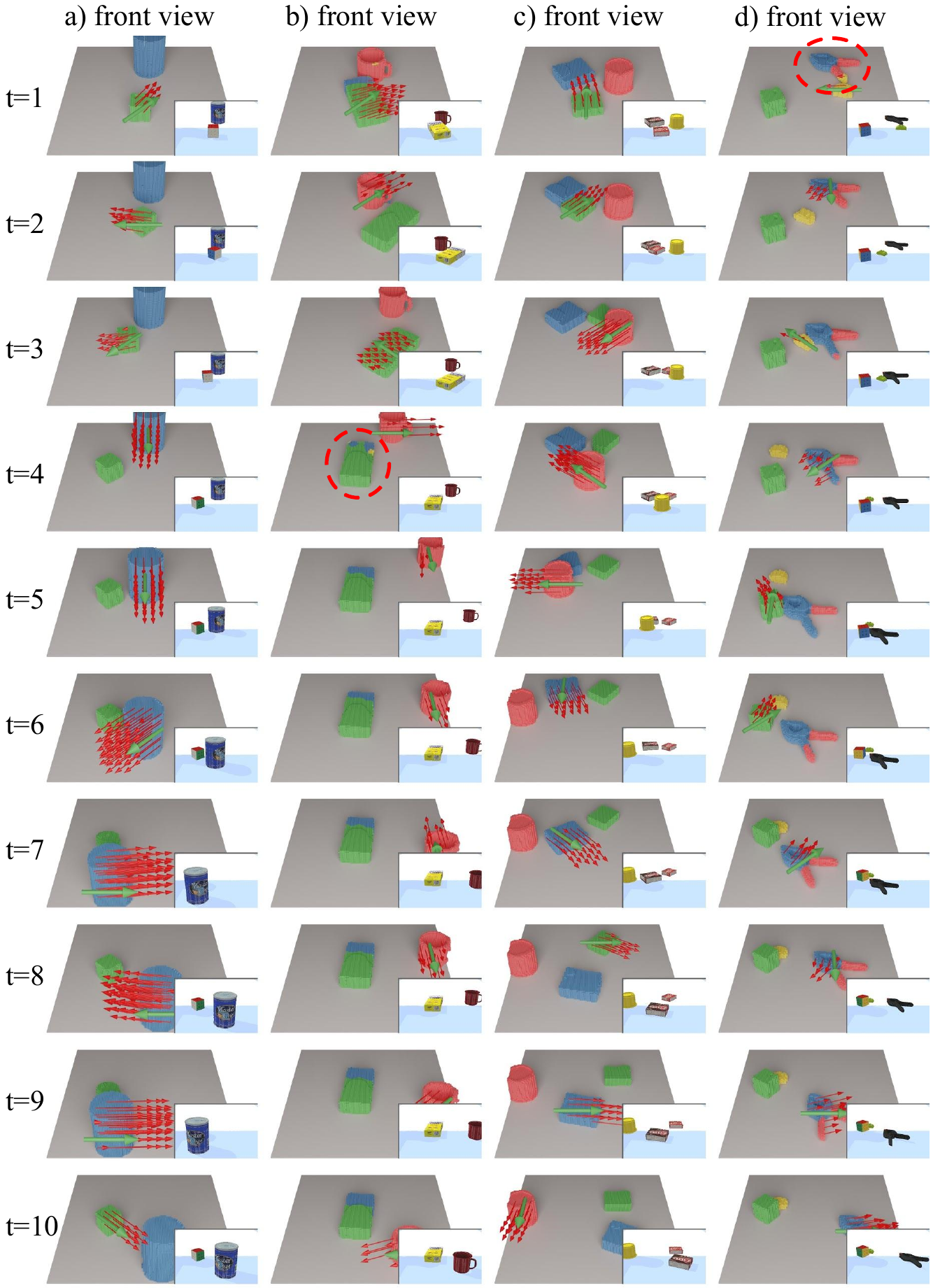}
    \caption{ \textbf{More test results on different number of objects.} We use four objects for training. During testing, we use test two objects (a, b) and three objects (c, d), without finetuning. While the algorithm is trained only on 4 object cases, it is able handle test case with fewer objects by predicting empty mask for additional object channels. Failure cases (b, d): If there are much noise in observation (uneven surface of sugarbox) or the object shape is different from training objects (clamp), the object may be mistakenly divided into two parts.}
    \label{fig:more_result-object_num}
    % https://docs.google.com/drawings/d/12uCLR_A4AwFW2_yvBMLmSv95rlTOwCnO7bIYvTy5Ln4/edit?usp=sharing
\end{figure}

Figure \ref{fig:more_result-object_num} shows additional result for tests cases with different number of objects (two or three object) compare to training cases (four object). While the algorithm is trained only on 4 object cases, it is able handle test case with fewer objects by predicting empty mask for additional object channels.

\section{Additional Results and Failure cases}
\begin{figure}[t]
    \centering
    \includegraphics[width=\linewidth]{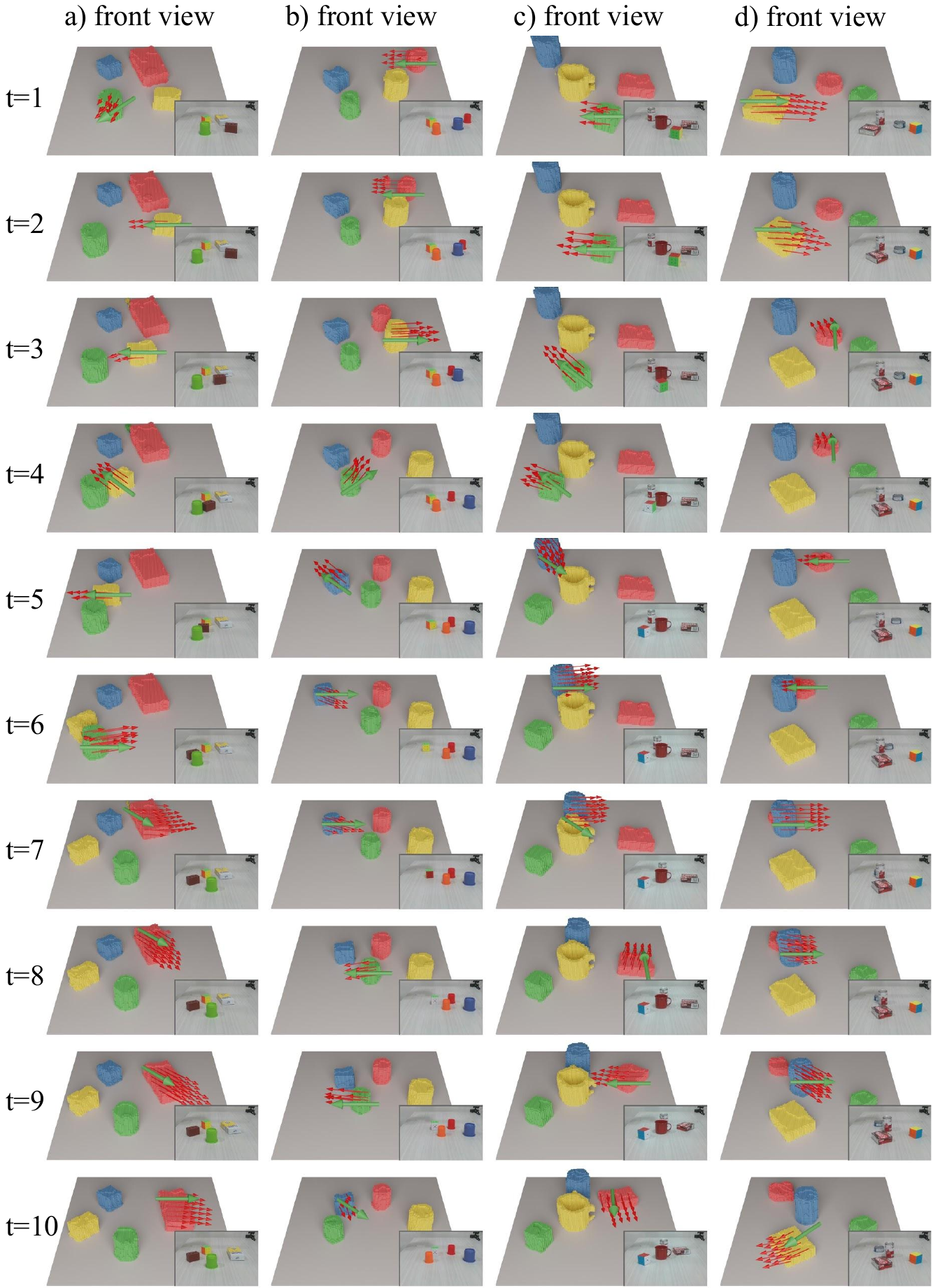}
    \caption{ \textbf{More test results on realworld dataset.} Object amodal instances mask are visualized in different colors. Image observation is shown in the bottom-right corner. Pushing action and predicted motion are represented by green and red arrows respectively.}
    \label{fig:more_result-success_1}
    % https://docs.google.com/drawings/d/1H_hzEzxdk7v69cAuC5Zj9pEl1goVX3KAEo2HsL8U32s/edit?usp=sharing
\end{figure}

\begin{figure}[t]
    \centering
    \includegraphics[width=\linewidth]{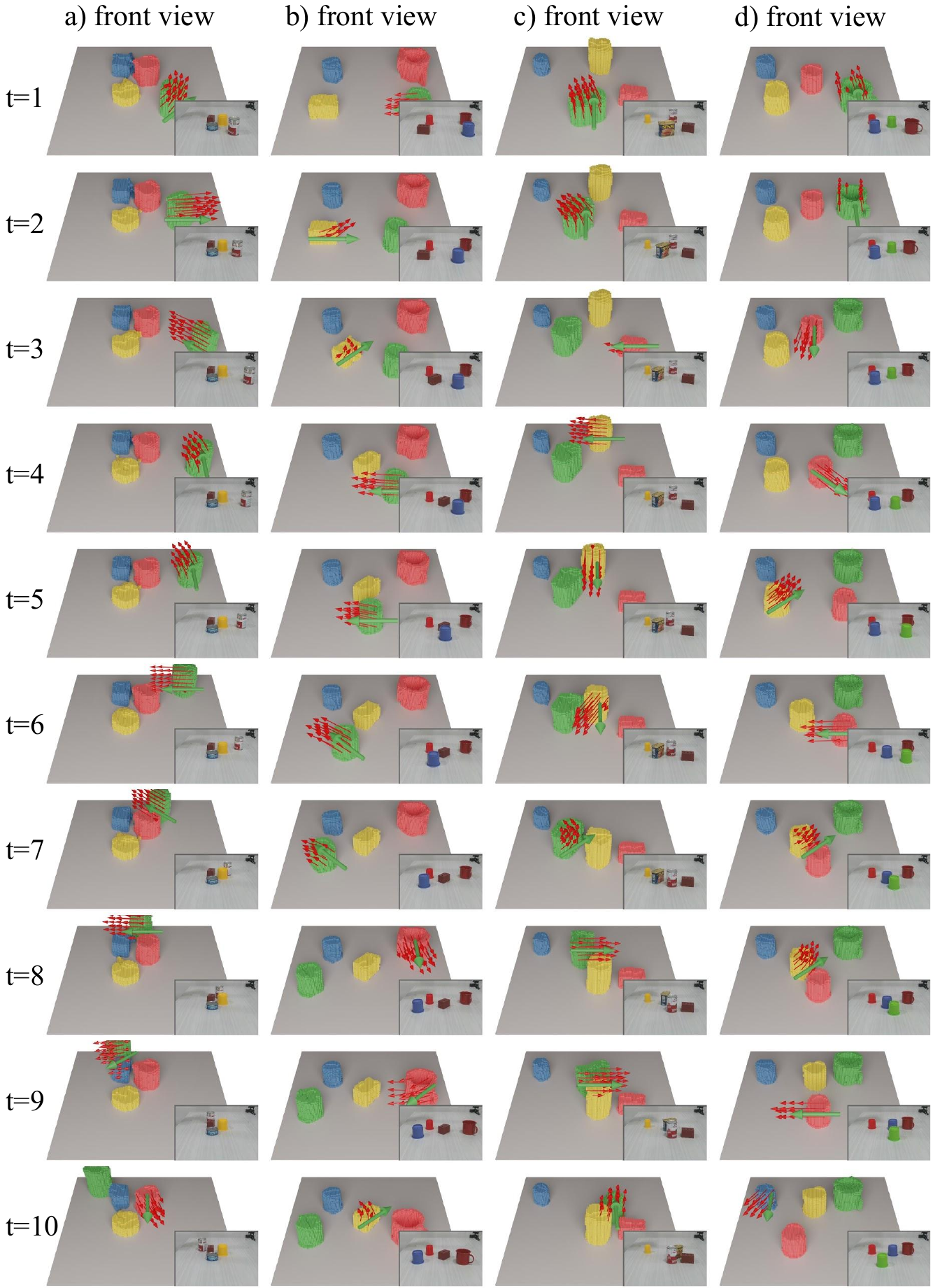}
    \caption{ \textbf{More test results on realworld dataset} (continue).}
    \label{fig:more_result-success_2}
    % https://docs.google.com/drawings/d/12wCfHwj8sB8eAJ_TL2emceJEbE6Fghaf6y0kC0iY4XI/edit?usp=sharing
\end{figure}

\begin{figure}[t]
    \centering
    \includegraphics[width=\linewidth]{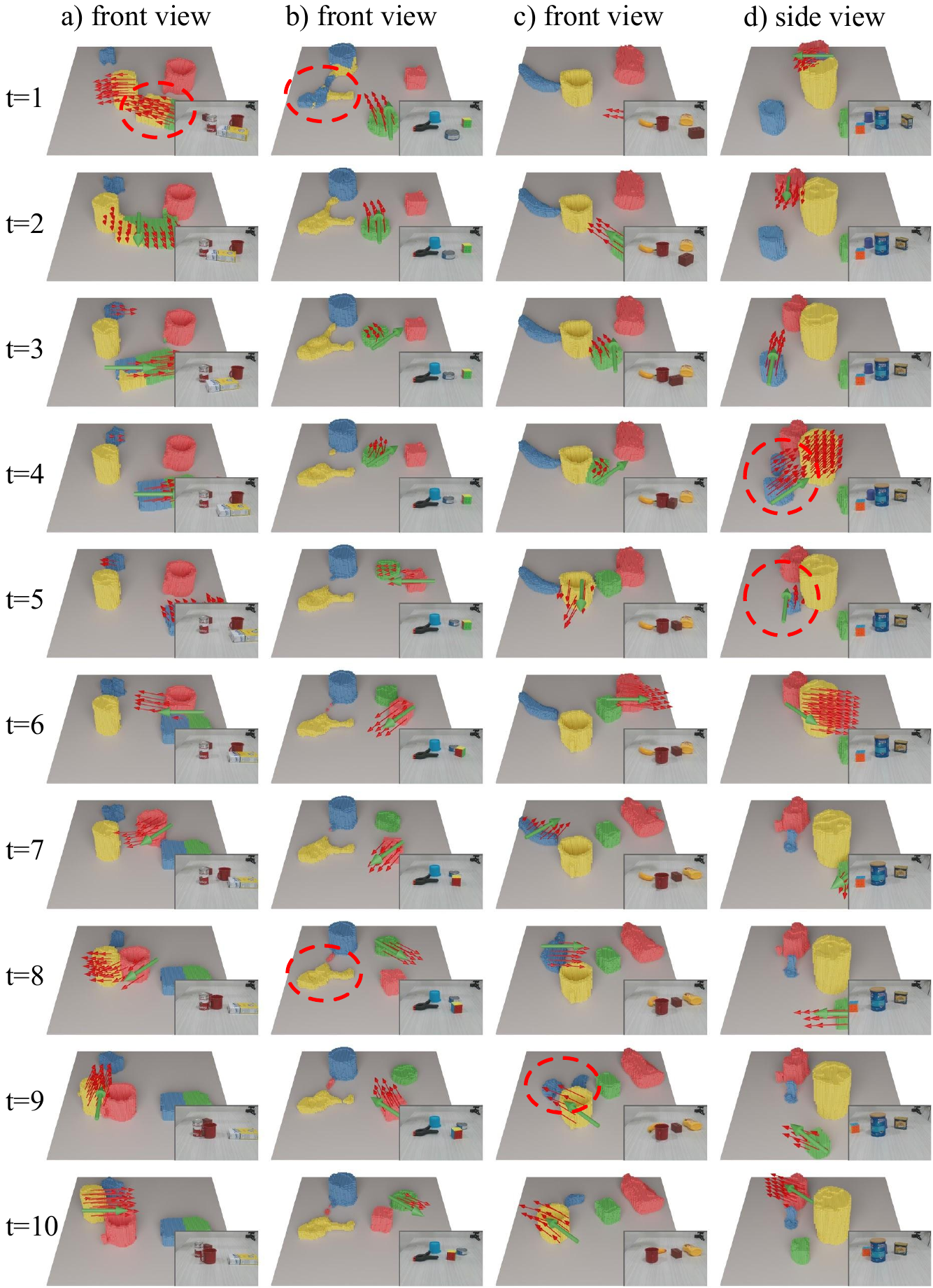}
    \caption{ \textbf{More test results on realworld dataset} (failure case). Typical failure steps are circled in red for each sequence. (a, b): The object is mistakenly divided into two parts. (c, d): The mask of occluded object is wrong due to inaccurate motion prediction and long-time occlusion.}
    \label{fig:more_result-failure}
    % https://docs.google.com/drawings/d/1Yc8ySE2G8u7vz2aOd9_3AUDcHSu9qLaMYEk6kDQbOPk/edit?usp=sharing
\end{figure}

Figure \ref{fig:more_result-success_1} and figure \ref{fig:more_result-success_2} show additional qualitative results on the real-world benchmark. Figure \ref{fig:more_result-failure} shows some failure cases.

We use simple concatenation for history aggregation, potentially limiting the robustness in the face of incorrect history and long-term history. An interesting future direction might consider using sequences such as LSTM and GRU to handle noises in the history representation. 
Our algorithm is designed based on the assumption of rigid objects and use SE3 transformations for motion prediction. Future works might consider relaxing this assumption to model deformable or articulated objects. Finally, convolution operations are limited in modeling long-range relationships between objects (e.g., collision), which limits the capacity to model complex interaction and motion, where ideas from graph neural networks may be borrowed to improve this aspect of dynamics modeling.

\end{document}